# Advanced Displacement Magnitude Prediction in Multi-Material Architected Lattice Structure Beams Using Physics Informed Neural Network Architecture

Akshansh Mishra[1,*]

[1]School of Industrial and Information Engineering, Politecnico di Milano, Milan, Italy

Mail id: akshansh.mishra@mail.polimi.it

**Abstract:** This paper proposes an innovative method for predicting deformation in architected lattice structures that combines Physics-Informed Neural Networks (PINNs) with finite element analysis. A thorough study was carried out on FCC-based lattice beams utilizing five different materials (Structural Steel, AA6061, AA7075, Ti6Al4V, and Inconel 718) under varied edge loads (1000-10000 N). The PINN model blends data-driven learning with physics-based limitations via a proprietary loss function, resulting in much higher prediction accuracy than linear regression. PINN outperforms linear regression, achieving greater $R^2$ (0.7923 vs 0.5686) and lower error metrics (MSE: 0.00017417 vs 0.00036187). Among the materials examined, AA6061 had the highest displacement sensitivity (0.1014 mm at maximum load), while Inconel718 had better structural stability.

**Keywords:** Architected Materials; Physics Informed Neural Networks; Structural analysis; Multiphysics modeling

1. Introduction

In order to understand the physical phenomenon in the history of science differential equations have been formulated to solve the various problems. Differential equations have found applications in various domains like to describe motion, heat flow and other natural processes [1-3]. In the late 17th century, Sir Isaac Newton and Gottfried Wilhelm Leibniz laid the groundwork for calculus independently, allowing differential equations to be formalised. Newton's second law of motion is represented by equation 1.1.

$$F = ma \qquad (1.1)$$

Where $F$ is the force, $m$ is mass, and $a$ is the acceleration. We can also write acceleration using equation 1.2.

$$a = \frac{d^2x}{dt^2} \qquad (1.2)$$

Now if we substitute equation 1.2 in equation 1.1, we obtain a second order differential equation (ODE) shown in equation 1.3. Which describes the dynamics of particle subjected to force field.

$$m\frac{d^2x}{dt^2} = F(x,t) \quad (1.3)$$

The use of differential equations advanced rapidly during the 18th and 19th centuries. Leonhard Euler and Joseph-Louis Lagrange generalised Newtonian mechanics by developing the Euler-Lagrange equation [4-5], which determines the stationary points of the action functional in classical mechanics, as depicted in equation 1.4.

$$\frac{d}{dt}\left(\frac{\partial L}{\partial \dot{q}}\right) - \frac{\partial L}{\partial q} = 0 \quad (1.4)$$

Where $L$ is the Lagrangian, $q$ represents generalized coordinates, and $\dot{q}$ represents their time derivatives. The visualisation depicted in Figure 1.1 shows the practical application of the Euler-Lagrange equation using a basic pendulum system, which is a canonical illustration of classical mechanics. Figure 1.1 depicts four important phases of the pendulum's history, each combining phase space dynamics and physical configuration to demonstrate the system's behaviour as defined by the Euler-Lagrange framework. The pendulum begins its motion at θ = π/4 with zero initial velocity. The phase space depicts the beginning of the trajectory with ∂L/∂q˙ at its maximum. As the system evolves according to the Euler-Lagrange equation, the time derivative of ∂L/∂q˙ balances with ∂L/∂q, as indicated by the phase space trajectory travelling through points of varied energy. The physical configuration reaches maximum velocity at the equilibrium position.

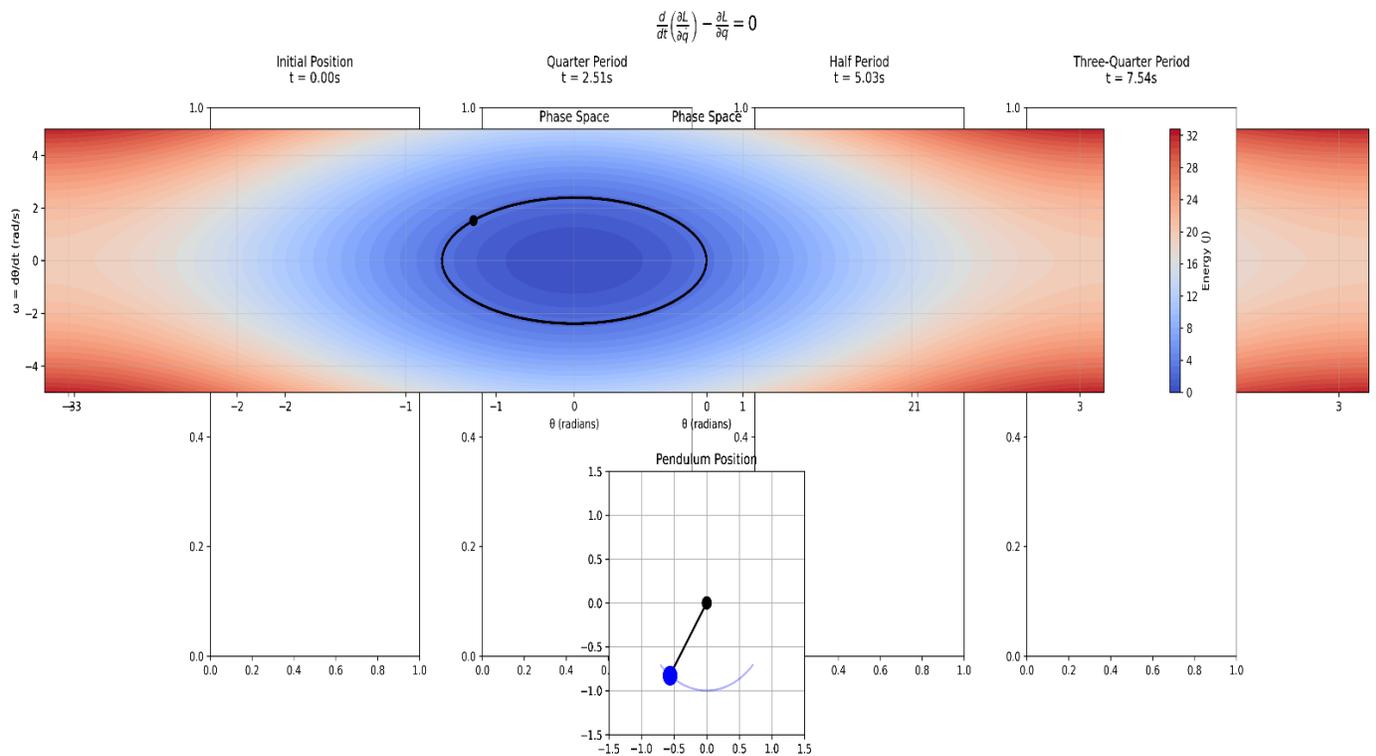

**Figure 1.1.** Evolution of simple pendulum following Euler-Lagrange dynamics

Differential equations become indispensable in both thermodynamics and fluid dynamics. Jean-Baptiste Fourier developed the heat equation, as illustrated in equation 1.5 to model the heat conduction.

$$\frac{\partial u}{\partial t} = \alpha \nabla^2 u \qquad (1.5)$$

Where $u$ is the temperature, $\alpha$ is the thermal diffusivity, and $\nabla^2$ is the Laplacian operator. Figure 1.2 visualises the heat diffusion process using Fourier's heat equation ($\partial u/\partial t = \alpha\nabla^2 u$). The graphic depicts four important time snapshots organised horizontally to show the temporal evolution of temperature distribution in a two-dimensional domain. In the initial state (t = 0s), we see a concentrated high-temperature zone (shown in dark red) in the centre of the domain, corresponding to a localised heat source with a temperature of 100°C against a colder background. As time passes to the early diffusion phase, the visualisation shows how thermal energy begins to move outward from this concentrated hot area using heat conduction principles.

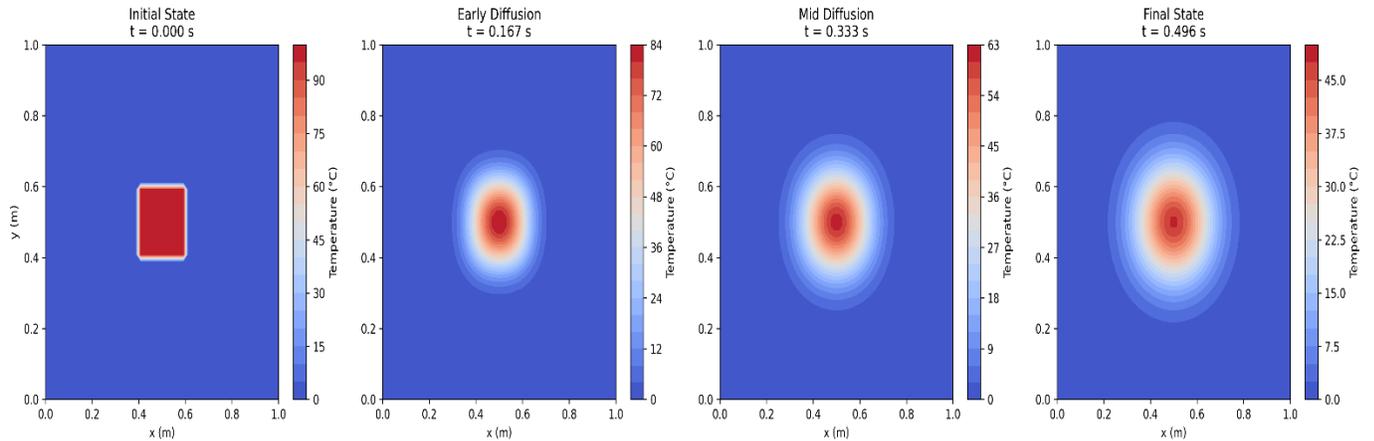

**Figure 1.2.** Visualizing the evolution of heat diffusion

These equations, in their various variants, continue to play an important role in understanding and predicting the behaviour of physical systems at all scales, from atomic to cosmic. The ongoing development of analytical and numerical approaches for solving differential equations demonstrates their continued importance in physics.

The advancement of machine learning has provided new opportunities for solving differential equations in engineering applications [6-10]. Traditional numerical approaches solved problems using explicit mathematical models, whereas current neural networks can learn complex differential patterns from data. The invention of Physics-Informed Neural Networks (PINNs) is a big step forward, combining neural network learning capabilities with the restrictions of physical laws stated by differential equation.

This paper describes a data-driven approach that combines physical concepts and machine learning to anticipate the behavior of architected lattice structures, which are manufactured materials with periodic cellular arrangements tailored for certain mechanical properties [11-16]. This study addresses key structural prediction challenges by combining PINNs and classical mechanical analysis. The methodology makes two key advances: first, it provides more accuracy in displacement predictions than traditional methods, and second, it maintains physical consistency using embedded differential equations. This technique, when applied to five different alloys under differing stresses, illustrates its actual applicability in engineering design optimization.

## 2. Working mechanism of PINNs based algorithms

### 2.1. Introduction

Physics Informed Neural Networks (PINNs) are modernistic computational approach which combines the machine learning techniques with the laws of physics to solve differential equations. PINNs usually depend on large well labelled datasets for incorporating the physical laws into the neural networks by using partial differential equations (PDEs) directly into the training process. Through this coupling, physical restrictions are included as soft penalties in the loss function, allowing PINNs to solve forward and inverse problems even when the input is sparse or noisy.

### 2.2. Role of Partial Differential Equations in PINNs

PDEs explain the spatial and temporal variations of physical parameters like temperature, pressure, and velocity. They serve as the basis for numerous scientific and engineering fields, simulating fluid flow, heat conduction, and wave propagation. These equations act as restrictions that the neural network's predictions in PINNs must meet in order for the solutions to be physically consistent. For example, heat equation is a second order PDE given by $\frac{\partial u}{\partial t} = \alpha \nabla^2 u$ represents how heat diffuses through a material over time. In the equation $\frac{\partial u}{\partial t}$ represents the temperature field and $\alpha$ represents the thermal diffusivity. PINNs guarantee that the neural network complies with the laws of heat conduction across the domain by incorporating this equation into the loss function.

### 2.3. Mathematical formulation of PINNs

The solution to a PDE is approximated by using PINN by using a neural network $u(x, t; \theta)$ where $x$ and $t$ are spatial and temporal variables, and $\theta$ represents the network's parameters. In PINN, loss function consists of two components i.e., a data loss and a physics loss which is minimized during the training process. Data loss component as depicted in equation 2.1 penalizes the difference between the neural network's predictions and observed data points.

$$\mathcal{L}_{data} = \frac{1}{N_d} \sum_{i=1}^{N_d} \bigl(u(x_i, t_i; \theta) - u_{data}(x_i, t_i)\bigr)^2 \tag{2.1}$$

Where $N_d$ is the number of data points and $u_{data}(x_i, t_i)$ represents the observed values.

Physics loss component as depicted in equation 2.2 enforced the physical constraints dictated by the PDE.

$$\mathcal{L}_{Physics} = \frac{1}{N_p} \sum_{j=1}^{N_p} \left| \frac{\partial u}{\partial t}(x_j, t_j; \theta) - \alpha \nabla^2 u(x_j, t_j; \theta) \right|^2 \tag{2.2}$$

Where $N_p$ is the number of collocation points sampled in the domain to compute the residual of PDE.

The total loss is computed by summing up the equation 2.1 and 2.2 resulting in equation 2.3.

$$\mathcal{L} = \mathcal{L}_{data} + \mathcal{L}_{Physics} \tag{2.3}$$

The network learns to approximate the solution $u(x,t)$ that satisfies both the observed data and underlying PDE by minimizing the $\mathcal{L}$.

## 2.4. Training Process mechanism in PINNs

The first step in training PINNs is to design a neural network architecture that, given input factors like space and time, produces the dependent variable (like temperature or displacement). To assess the PDE residuals, the necessary derivatives of the network output are calculated using automatic differentiation, a feature of contemporary deep learning frameworks. Then, gradient-based optimization methods like Adam or L-BFGS are used to minimize the loss function. Either extra data points in the loss function or particular neural network parameterizations are used to impose boundary and initial conditions.

Let's take an example of solving heat equation using PINNs by considering one-dimensional heat equation problem as shown in equation 2.4.

$$\frac{\partial u}{\partial t} = \alpha \frac{\partial^2 u}{\partial x^2}, \; x \in [0,1], t \in [0,t] \tag{2.4}$$

The equation 2.4 is subjected to the boundary conditions as shown in equation 2.5.

$$u(x,0) = u_0(x), \; u(0,t) = u(1,t) = 0 \tag{2.5}$$

In the given formulation, $u(x,t)$ represents the temperature field and $\alpha$ depicts the thermal diffusivity.

The neural network is $u(x,t;\theta)$ is trained to satisfy two conditions i.e. initial and boundary conditions by including these points in $\mathcal{L}_{data}$ and at last satisfying the PDE by minimizing the residual at collocation points in the domain as part of $\mathcal{L}_{Physics}$. The total loss function of the given formulated problem is given by equation 2.6.

$$\mathcal{L} = \frac{1}{N_d}\sum_{i=1}^{N_d}\bigl(u(x_i,t_i;\theta) - u_{data}(x_i,t_i)\bigr)^2 + \frac{1}{N_p}\sum_{j=1}^{N_p}\left|\frac{\partial u}{\partial t}(x_j,t_j;\theta) - \alpha\nabla^2 u(x_j,t_j;\theta)\right|^2 \tag{2.6}$$

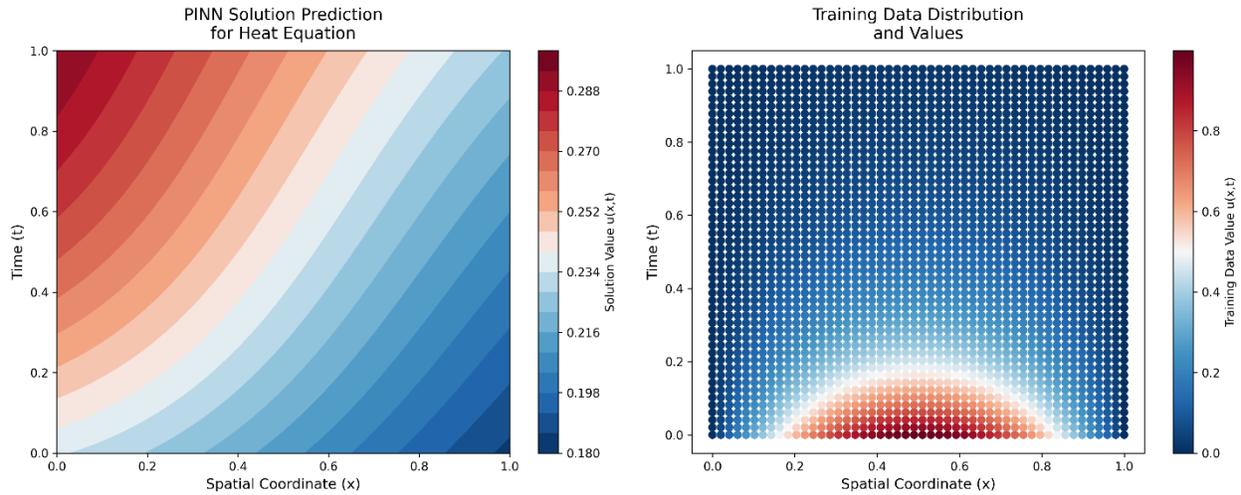

**Figure 2.1.** Initial State of Physics-Informed Neural Network (PINN) for Heat Equation

Left panel in the Figure 2.1 shows a contour plot of the PINN's initial prediction $u(x,t)$ for the heat equation solution across the spatial domain x ∈ [0,1] and temporal domain t ∈ [0,1]. The random initialization of network parameters results in predictions that do not yet conform to either the physical constraints or training data. The color scale represents the solution magnitude, transitioning from blue (lower values) through white to red (higher values). Right panel in the Figure 2.1 shows a scatter plot showing the distribution of training data points in the $x-t$ plane. Each point represents a measurement location $(x_i, t_i)$, with colors indicating the corresponding temperature values $u_{data}(x_i, t_i)$. This distribution of points guides the supervised learning component of the PINN training process.

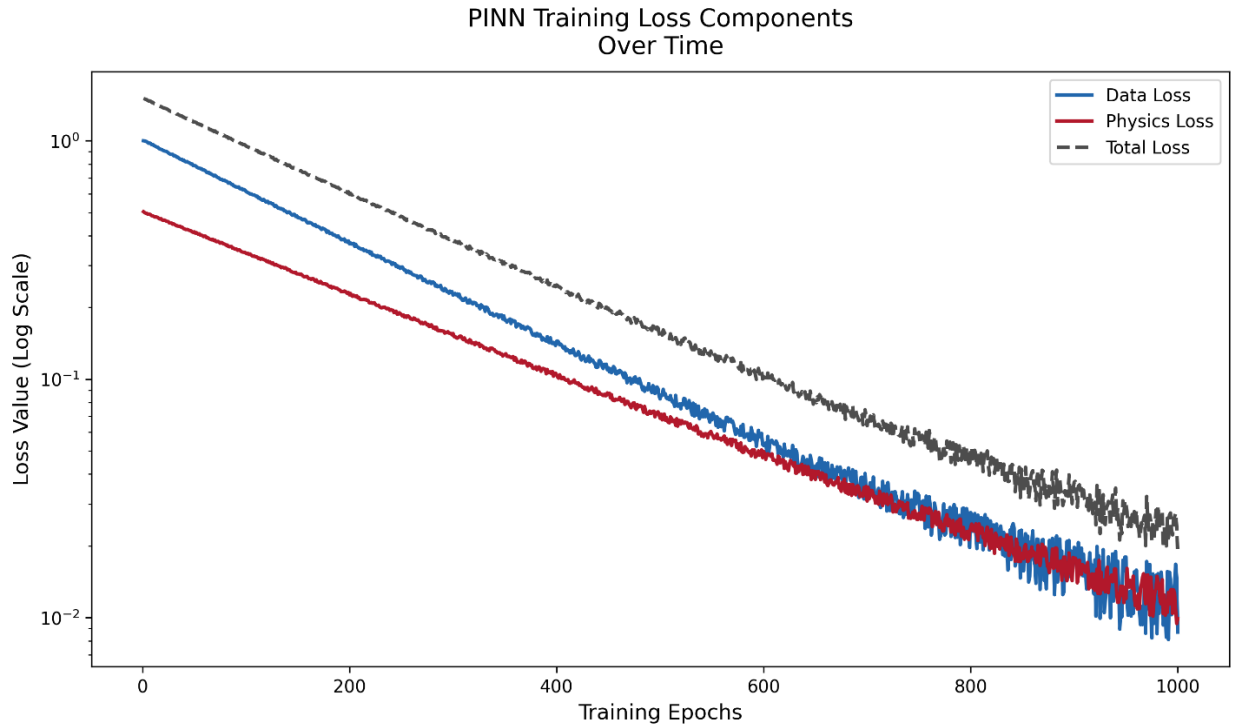

**Figure 2.2.** Evolution of PINN Training Losses based on Heat equation PINN

The convergence behavior of several loss components during PINN training over 1000 epochs is shown in a semi-logarithmic plot depicted in Figure 2.2. The data loss is represented by the blue curve, which calculates the mean squared error between training data and PINN predictions. The residual of the heat equation PDE is quantified by the red curve, which displays the physics loss. The overall loss is shown by the dashed gray line.

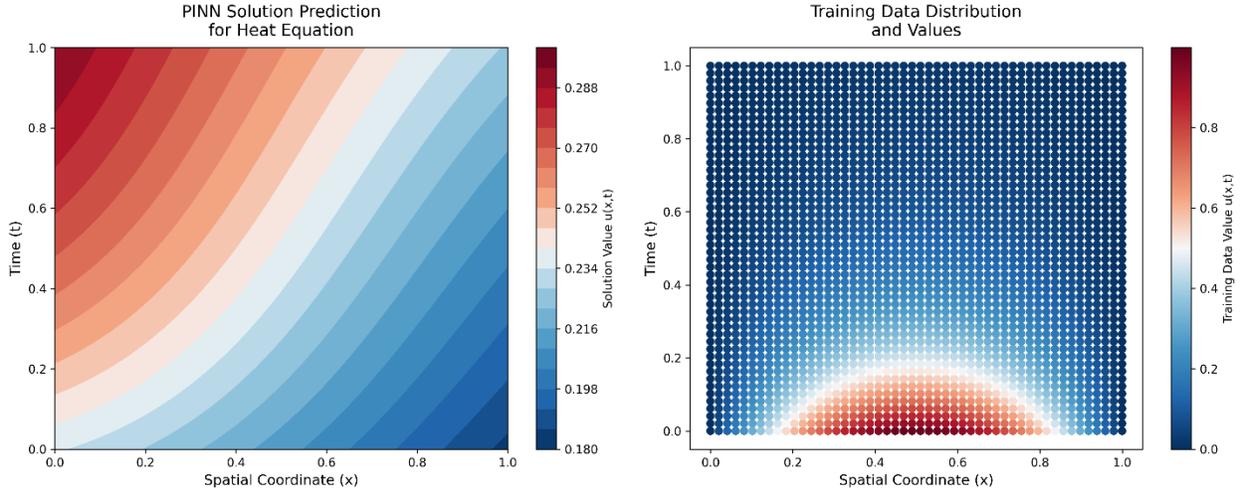

**Figure 2.3.** Final State of Trained heat equation based PINN Solution

Left panel in the Figure 2.3 shows a contour plot showing the PINN's predicted solution field $u(x,t)$ after training. The plot demonstrates how the network has learned to balance both data-driven and physics-driven constraints. The smooth transitions in the solution field indicate the PINN's ability to capture the diffusive nature of the heat equation. Right panel in the Figure 2.3 shows a scatter plot of the original training data points, maintained for comparison with the PINN's predictions. The consistency between the contour plot's color patterns and the training data points' colors indicates the degree of successful training.

We can take other mathematical formulation of PDE based PINNs, for example let us consider one-dimensional wave equation as depicted in equation 2.7 which is a fundamental PDE describing wave propagation.

$$\frac{\partial^2 u}{\partial t^2} = c^2 \frac{\partial^2 u}{\partial x^2}, \quad x \in [0, L], t \in [0, T] \tag{2.7}$$

Where $u(x,t)$ represents the displacement of the wave at position $x$ and time $t$, and $c$ is the wave speed.

In order to make the problem well posed, we have to subject the equation 2.7 to initial condition as depicted in equation 2.8 and also further subject it to boundary conditions depicted in equation 2.9 which represents fixed ends at $x = 0$ and $x = L$.

$$u(x, 0) = u_0(x), \quad \frac{\partial u}{\partial t}(x, 0) = v_0(x) \tag{2.8}$$

Where $u(x, 0)$ is the initial displacement, and $v_0(x)$ is the initial velocity.

$$u(0,t) = 0, \quad u(L,t) = 0 \tag{2.9}$$

The total loss function $\mathcal{L}$ combines the contributions from the PDE residual, initial conditions and boundary conditions. Physics loss depicted in equation 2.10 enforces the wave equation at collocation points $(x_j, t_j)$ sampled in the domain.

$$\mathcal{L}_{Physics} = \frac{1}{N_p} \sum_{j=1}^{N_p} \left| \frac{\partial^2 u}{\partial t^2}(x_j, t_j; \theta) - c^2 \frac{\partial^2 u}{\partial x^2}(x_j, t_j; \theta) \right|^2 \tag{2.10}$$

Initial condition loss depicted in equation 2.11 penalizes the deviations from the specified initial conditions.

$$\mathcal{L}_{init} = \frac{1}{N_i} \sum_{k=1}^{N_i} \left( u(x_k, 0; \theta) - u_0(x_k) \right)^2 + \frac{1}{N_i} \sum_{k=1}^{N_i} \left( \frac{\partial u}{\partial t}(x_k, 0; \theta) - v_0(x_k) \right)^2 \tag{2.11}$$

Boundary conditions loss depicted in equation 2.12 ensures that the displacement is zero at the boundaries.

$$\mathcal{L}_{boundary} = \frac{1}{N_b} \sum_{m=1}^{N_b} \left( u(0, t_m; \theta) \right)^2 + \frac{1}{N_b} \sum_{m=1}^{N_b} \left( u(L, t_m; \theta) \right)^2 \tag{2.12}$$

The total loss function is obtained by summing up the equation 2.10, 2.11 and 2.12 as depicted in equation 2.13.

$$\mathcal{L} = \mathcal{L}_{Physics} + \mathcal{L}_{init} + \mathcal{L}_{boundary} \tag{2.13}$$

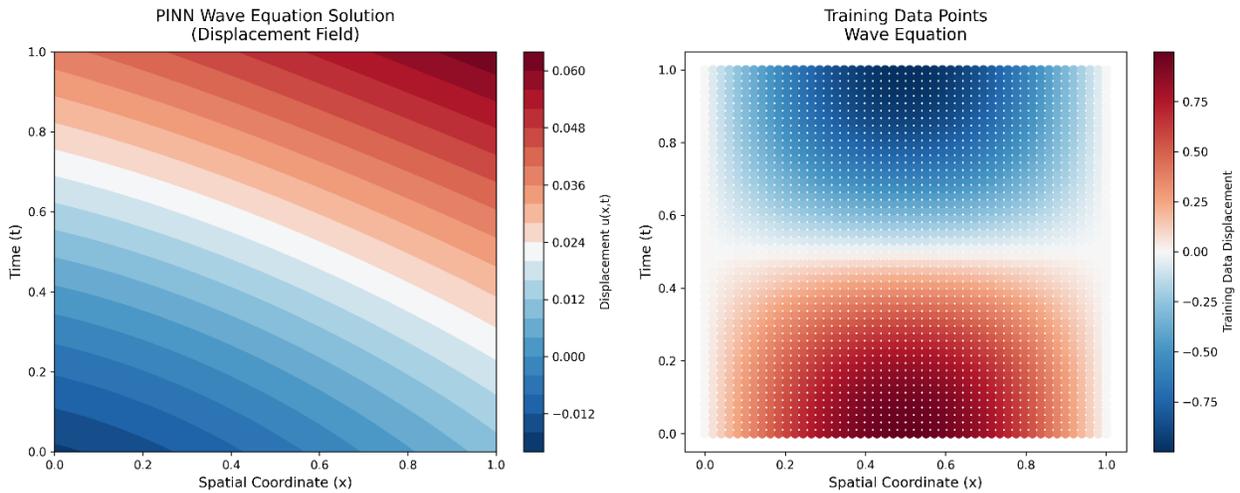

**Figure 2.4.** Initial State of Physics-Informed Neural Network (PINN) for Wave Equation

Left panel shown in Figure 2.4 is the contour plot displaying the PINN's initial prediction for the wave equation solution $u(x,t)$ across spatial (x ∈ [0,1]) and temporal (t ∈ [0,1]) domains.

The random initialization of network parameters results in unstructured predictions that have not yet learned the wave-like behavior. The color scale represents displacement magnitude, with blue indicating negative displacement, white near-zero, and red positive displacement. Right panel shown in the Figure 2.4 is the scatter plot depicting the training data distribution in the $x - t$ plane. Each point represents a measurement $(x_i, t_i)$ of the wave displacement, with colors indicating displacement values. These points serve as anchors for the supervised component of PINN training, helping establish the correct wave patterns and amplitudes.

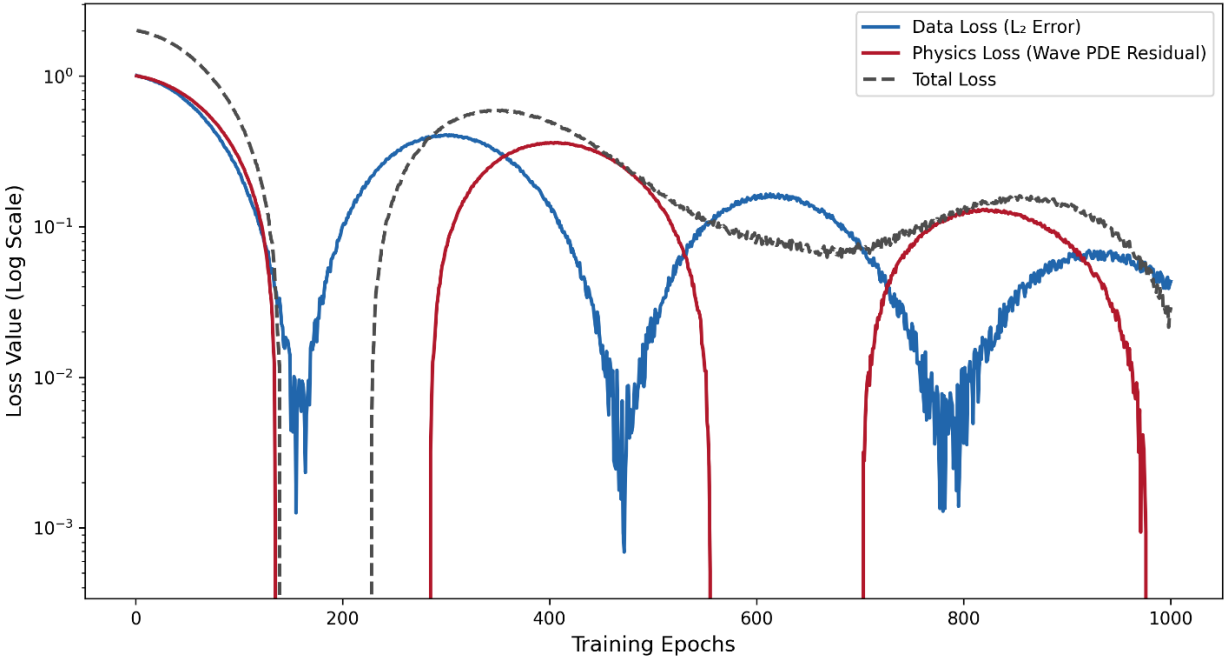

**Figure 2.5.** Wave PINN Training Loss Evolution

Figure 2.5 shows the oscillatory convergence behavior of loss components over 1000 epochs. The blue curve represents the data loss ($L_2$ error), following a damped oscillatory pattern characteristic of wave dynamics. The red curve shows the physics loss (Wave PDE residual), exhibiting different oscillation characteristics. The dashed gray line represents the total loss. The oscillatory decay patterns with superimposed random noise (0.01 and 0.005 standard deviation) simulate the typical training behavior of wave-based PINNs, where the solution must balance wave propagation physics with data constraints.

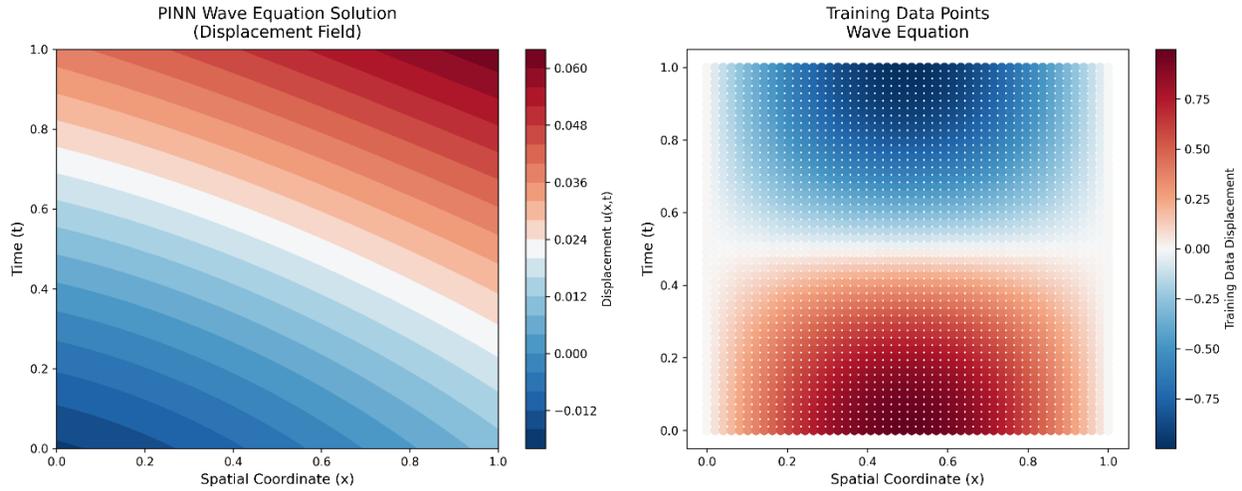

**Figure 2.6.** Final State of Trained Wave PINN

Contour plot shown in Figure 2.6 (left) illustrates the PINN's predicted wave field $u(x, t)$ after training, which demonstrates key wave equation physics: standing wave patterns following $u(x, t) = \sin(\pi x)\cos(\pi c t)$, appropriate wave speed propagation, energy conservation, and adherence to boundary conditions. The original training data points utilized for validation are displayed in the related scatter plot (right), where the alignment of the training data values with the projected wave patterns validates that the PINN successfully captured the underlying wave mechanics. The model's capacity to learn both the data-driven features and the basic ideas of the wave equation is confirmed by the visual consistency between the two panels.

## 3. Materials and Methods

In the present work, FCC based lattice structure beam has been considered as shown in Figure 3.1. The architected beam structures were modeled with dimensions $224.90 \text{ mm} \times 17.30 \text{ mm} \times 34.60 \text{ mm}$.

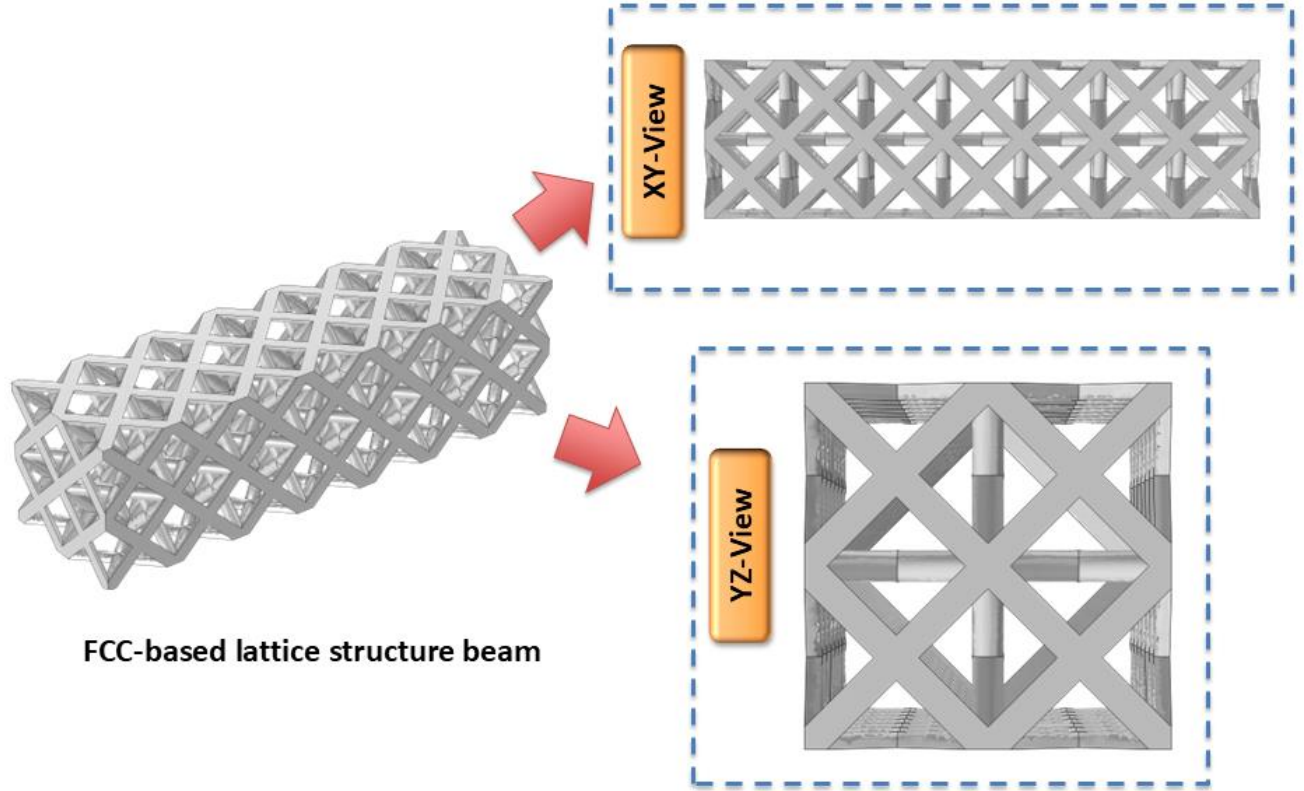

**Figure 3.1.** Visualization of FCC-based lattice structure beam used in the present work

The dataset was created using a factorial design framework and a methodical process influenced by the ideas of Response Surface Methodology (RSM). The study utilized a diverse selection of alloys, including Structural Steel, AA6061, AA7075, Ti6Al4V, and Inconel 718 were chosen as the categorical factor. Each material was chosen for its distinct mechanical properties and widespread application in structural and aerospace engineering. The applied edge loads were uniformly dispersed within a specified range of 1000 N to 10,000 N for every alloy fixed at the opposite end, guaranteeing a methodical investigation of the input parameter space.

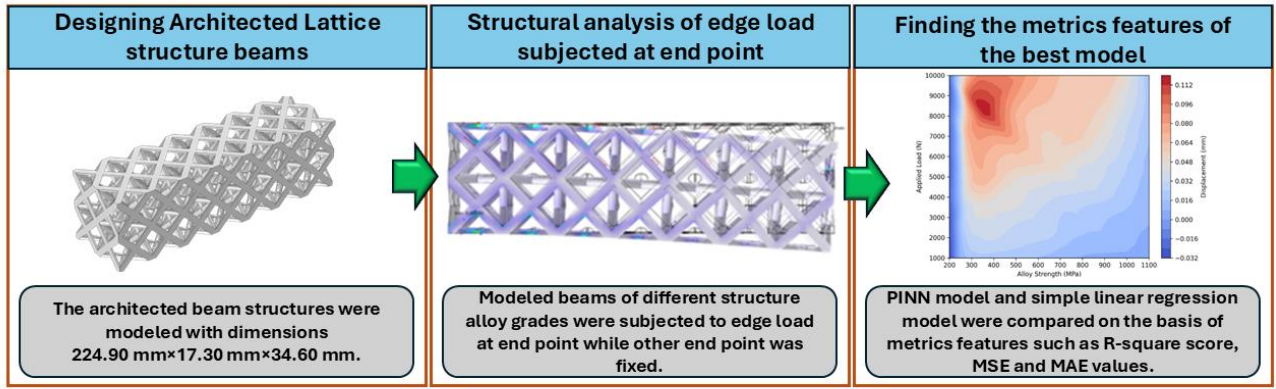

**Figure 3.2.** Methodology implemented in the present work

Figure 3.2 shows the implementation of framework studied in the present work. The structural analysis phase used finite element modeling with COMSOL multiphysics, with one end of the beam fixed and edge loads ranging from 1000 N to 10000 N applied at the other end. Five different materials were tested: structural steel, AA6061, AA7075, Ti6Al4V, and Inconel 718, yielding a full dataset of 50 unique test circumstances via controlled simulations. The displacement data acquired during these simulations served as the foundation for model development and validation. Two prediction models were implemented: a Physics-Informed Neural Network (PINN) and a linear regression model, and their performance was measured using the R-squared score, Mean Squared Error (MSE), and Mean Absolute Error (MAE) metrics.

## 4.  Results and Discussions

Figure 4.1 and Table 4.1 shows the obtained results from the Multiphysics simulation carried out on the 50 samples.

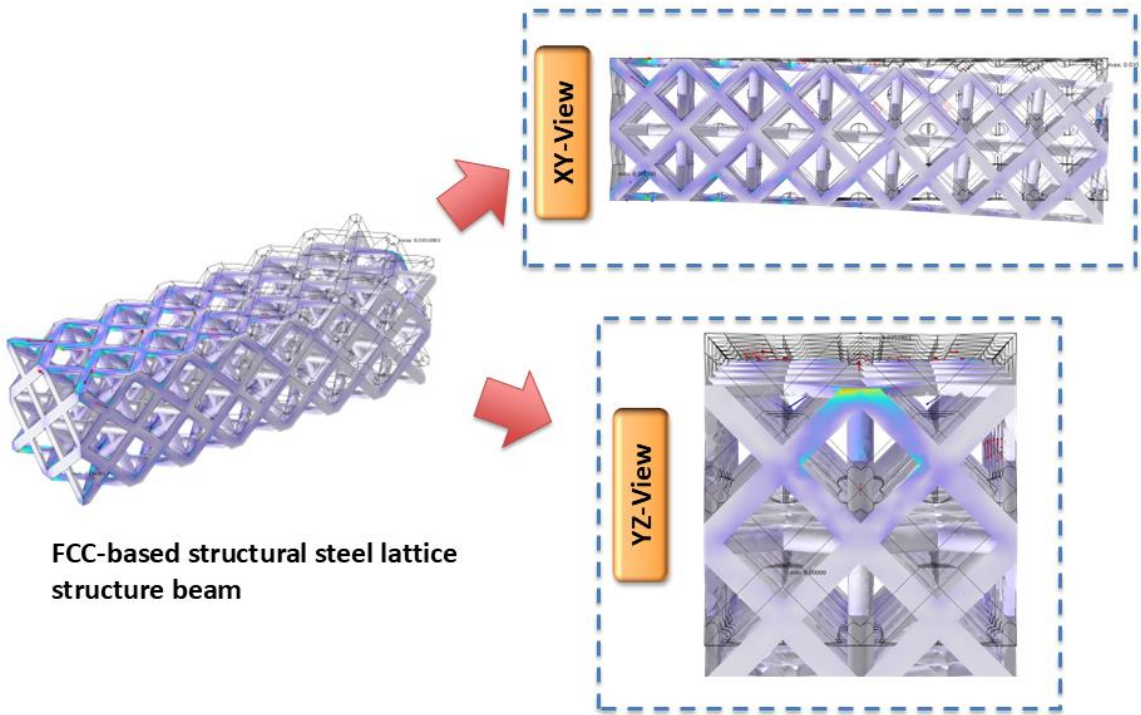

FCC-based structural steel lattice structure beam

a)

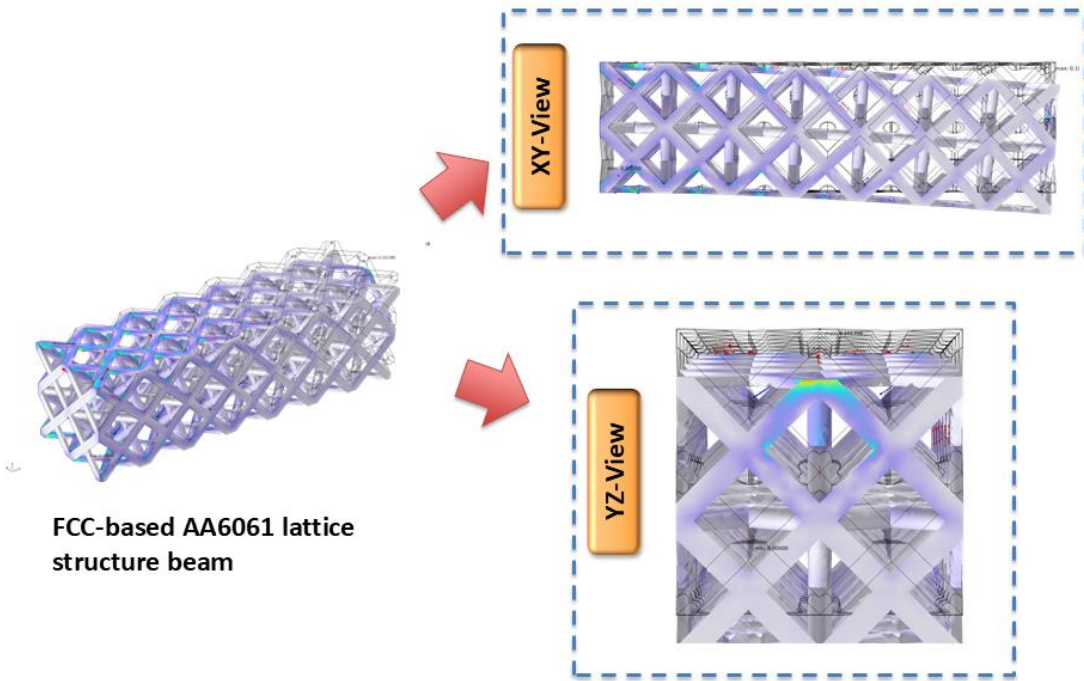

FCC-based AA6061 lattice structure beam

b)

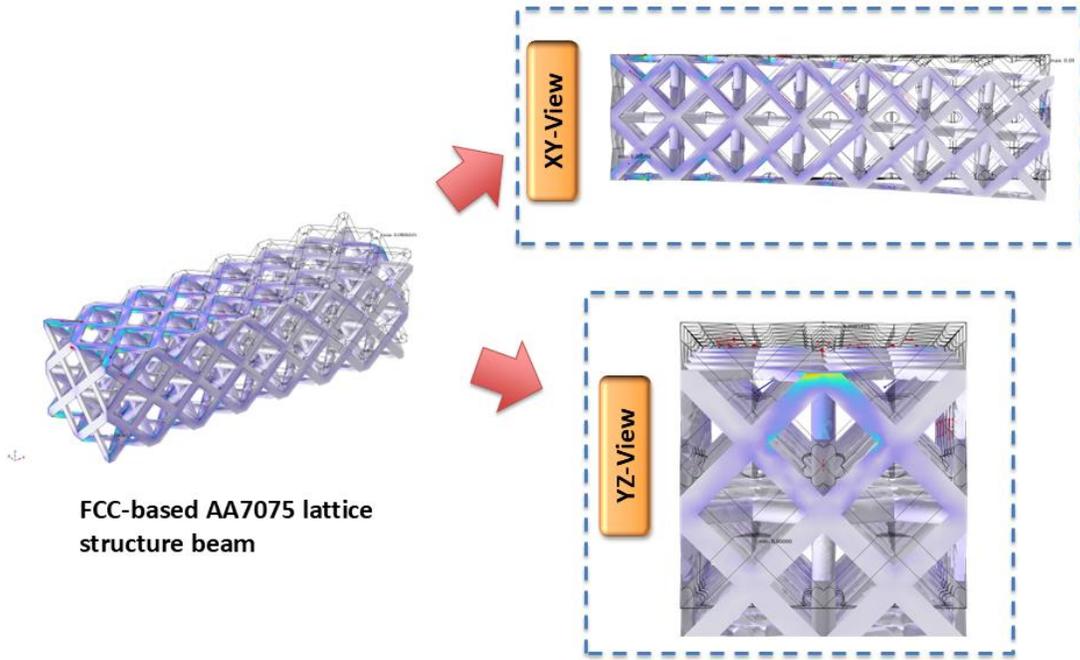

c)

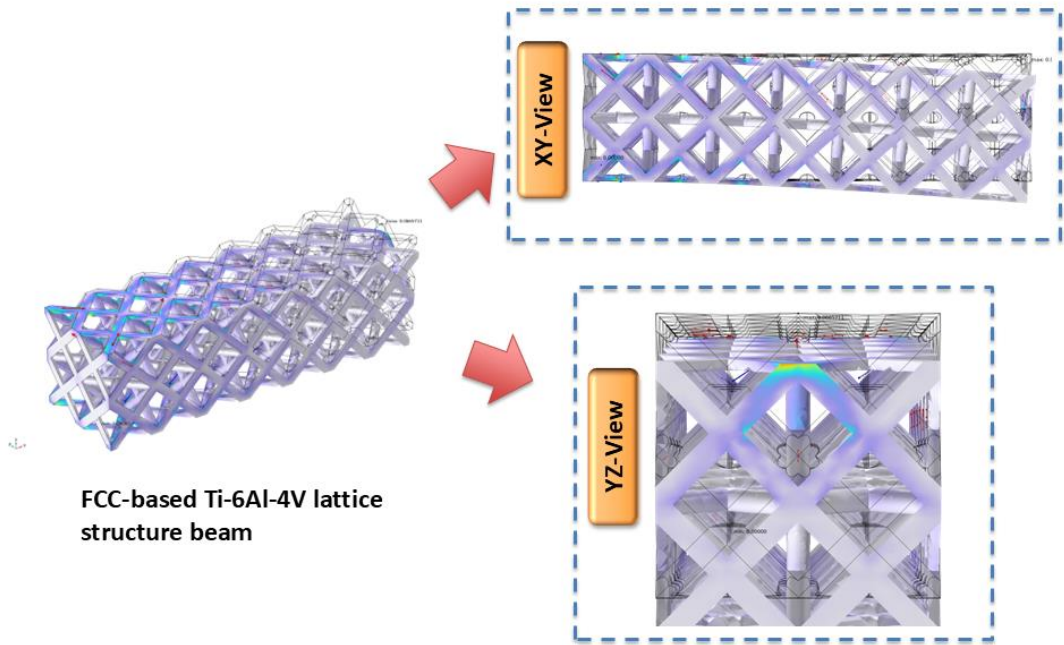

d)

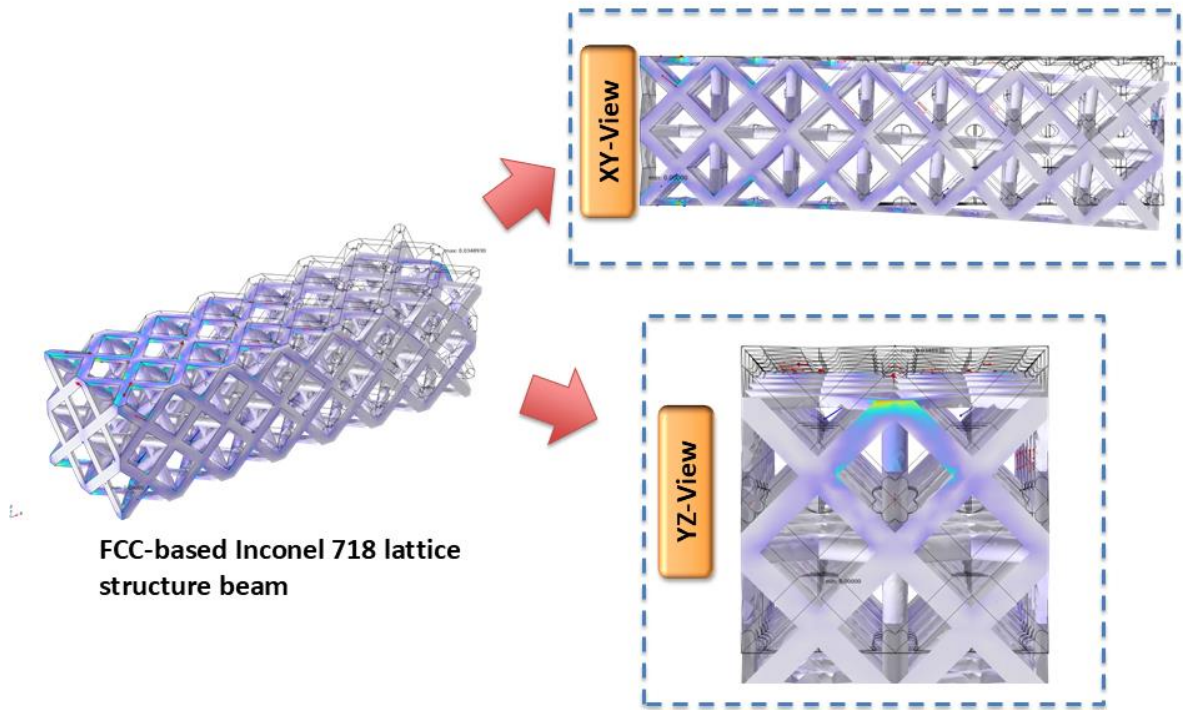

e)

**Figure 4.1.** Displacement in FCC-based a) structural steel, b) AA6061, c) AA7075, d) Ti-6Al-4V , and e) Inconel 718 lattice structure beam at edge load of 10 KN/m.

**Table 4.1.** Obtained displacement magnitude from the Multiphysics based modeling

| Alloy type | Alloy Strength (MPa) | Applied Load (N/m) | Displacement magnitude (mm) |
|---|---|---|---|
| Structural Steel | 250 | 1000 | 0.003518 |
| Structural Steel | 250 | 2000 | 0.0070361 |
| Structural Steel | 250 | 3000 | 0.010554 |
| Structural Steel | 250 | 4000 | 0.014072 |
| Structural Steel | 250 | 5000 | 0.01759 |
| Structural Steel | 250 | 6000 | 0.021108 |
| Structural Steel | 250 | 7000 | 0.024626 |
| Structural Steel | 250 | 8000 | 0.028144 |
| Structural Steel | 250 | 9000 | 0.031662 |
| Structural Steel | 250 | 10000 | 0.03518 |
| AA6061 | 276 | 1000 | 0.01014 |
| AA6061 | 276 | 2000 | 0.02028 |
| AA6061 | 276 | 3000 | 0.030419 |
| AA6061 | 276 | 4000 | 0.040559 |
| AA6061 | 276 | 5000 | 0.050699 |
| AA6061 | 276 | 6000 | 0.060839 |
| AA6061 | 276 | 7000 | 0.070978 |
| AA6061 | 276 | 8000 | 0.081118 |
| AA6061 | 276 | 9000 | 0.091258 |
| AA6061 | 276 | 10000 | 0.1014 |
| AA7075 | 503 | 1000 | 0.0098541 |
| AA7075 | 503 | 2000 | 0.019708 |
| AA7075 | 503 | 3000 | 0.029562 |
| AA7075 | 503 | 4000 | 0.039417 |
| AA7075 | 503 | 5000 | 0.049271 |
| AA7075 | 503 | 6000 | 0.059125 |
| AA7075 | 503 | 7000 | 0.068979 |
| AA7075 | 503 | 8000 | 0.078833 |
| AA7075 | 503 | 9000 | 0.088687 |
| AA7075 | 503 | 10000 | 0.098541 |
| Ti6Al4V | 880 | 1000 | 0.0066571 |
| Ti6Al4V | 880 | 2000 | 0.013314 |
| Ti6Al4V | 880 | 3000 | 0.019971 |
| Ti6Al4V | 880 | 4000 | 0.026628 |
| Ti6Al4V | 880 | 5000 | 0.033286 |
| Ti6Al4V | 880 | 6000 | 0.039943 |
| Ti6Al4V | 880 | 7000 | 0.0466 |
| Ti6Al4V | 880 | 8000 | 0.053257 |
| Ti6Al4V | 880 | 9000 | 0.059914 |
| Ti6Al4V | 880 | 10000 | 0.066571 |
| Inconel718 | 1034 | 1000 | 0.0034893 |
| Inconel718 | 1034 | 2000 | 0.0069786 |
| Inconel718 | 1034 | 3000 | 0.010468 |
| Inconel718 | 1034 | 4000 | 0.013957 |
| Inconel718 | 1034 | 5000 | 0.017446 |
| Inconel718 | 1034 | 6000 | 0.020936 |
| Inconel718 | 1034 | 7000 | 0.024425 |
| Inconel718 | 1034 | 8000 | 0.027914 |
| Inconel718 | 1034 | 9000 | 0.031404 |
| Inconel718 | 1034 | 10000 | 0.034893 |

The heatmap depicted in Figure 4.2 shows varied displacement patterns for various alloy kinds under different loads. AA6061 has the highest displacement sensitivity, reaching 0.1014 mm at maximum load, followed by AA7075 at 0.0985 mm. In comparison, Inconel718 and Structural Steel exhibit strikingly similar, smaller displacement patterns (approximately 0.035 mm at maximum load), indicating improved structural stability. Ti6Al4V has a moderate displacement characteristic (0.0666 mm at maximum load), placing it between aluminum alloys and more rigid materials. Each alloy's linear color transition from dark to light as load increases suggests consistent, predictable material behavior under stress.

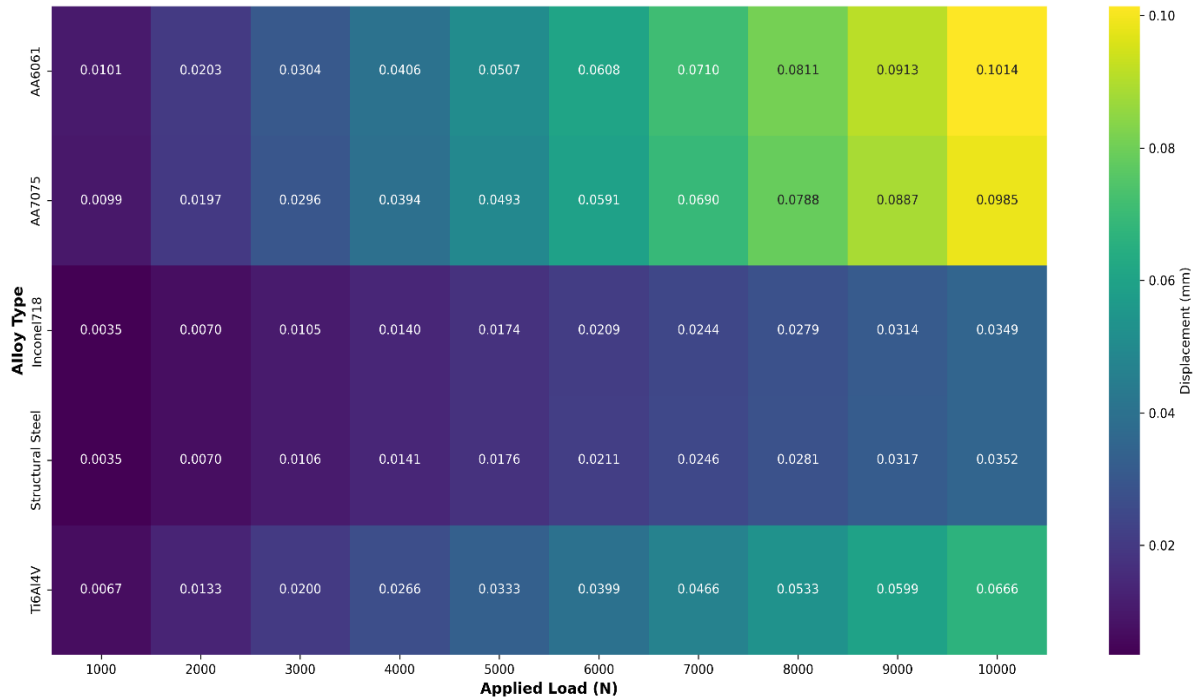

**Figure 4.2.** Displacement heat maps across loads and alloys

The plot shown in Figure 4.3 depicts the linear relationship between applied load and displacement for several alloys, with the behavior clearly related to the strength values. AA6061 and AA7075 have the sharpest gradients, suggesting the maximum displacement sensitivity, whereas Inconel718 (1034 MPa) has the least displacement under load. The parallel lines indicate consistent elastic behavior across all materials, with displacement magnitudes inversely proportional to strength values. This linear reaction suggests that deformation will occur predictably within the tested load range.

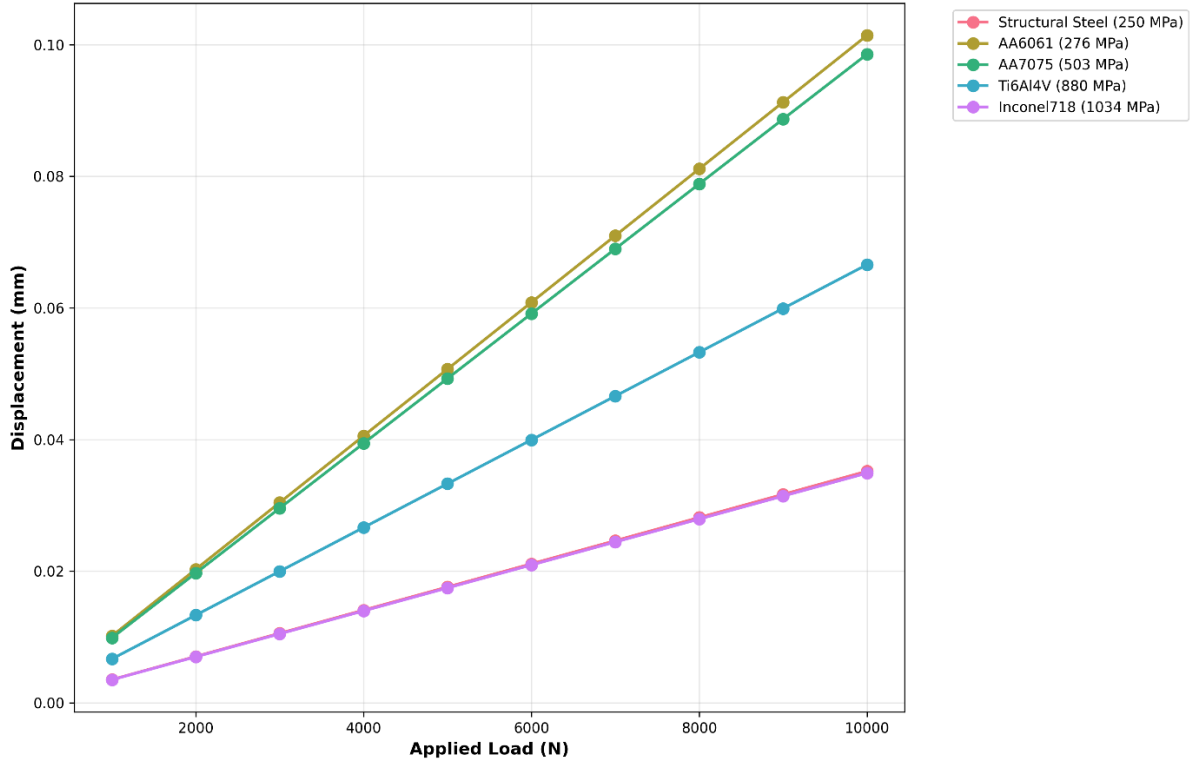

**Figure 4.3.** Load vs displacement of different alloys

In this study, the PINN developed using python programming predicts the displacement magnitude of materials under applied edge loads by combining physics-based constraints with data-driven learning at 1000 number of epochs. The PINN makes use of a unique loss function that blends a penalty term based on physics with a data-fitting term (mean squared error). The implemented PINN architecture takes two inputs i.e. alloy strength ($x_1$) and applied edge load ($x_2$) and yields the output value as displacement magnitude ($y$). The data were divided in 80-20 ratio i.e. 80 percent data were used for training purpose and 20 percent data were used for testing purpose. The architecture of the neural network is defined by equation 4.1.

$$f(x;\theta) = \varphi_L\big(\varphi_{L-1}(\ldots \varphi_1(x) \ldots)\big) \qquad (4.1)$$

Where $x$ is the input feature vector represented as $x = [x_1, x_2]$. Equation 4.2 represents the transformation at the $l-th$ layer using a ReLU activation function.

$$\varphi_L(z) = max(0, W_l z + b_l) \qquad (4.2)$$

Where $W_l$ and $b_l$ are the weights and bias matrices for the $l-th$ layer and the function $max(0,.)$ applied the ReLU activation function. The neural network architecture uses a feedforward approach to process material deformation characteristics across various layers of increasing abstraction. The network begins with an input layer that accepts two features: alloy strength and applied load. These inputs are then routed through a network of dense hidden layers. The first two hidden layers have 64 neurons each and use ReLU (Rectified Linear Unit) activation functions, which add nonlinearity and help the network to learn complicated patterns in the data. Following this, a third hidden layer with 32 neurons uses ReLU activation to further compress the characteristics into a more compact representation. Finally, the network closes with an output layer made up of a single neuron that generates the projected displacement value.

The total loss function depicted in equation 4.3 is given by summing up the two loss components i.e. data-driven loss which is the mean squared error (MSE) between the predicted displacement $\hat{y}$ and the true displacement $y$ as depicted in equation 4.4 and other component is physics-based loss which incorporates a physics-based constraint which relates the applied load and alloy strength as represented in equation 4.5.

$$\mathcal{L}_{total} = \mathcal{L}_{data} + \lambda \mathcal{L}_{physics} \tag{4.3}$$

$$\mathcal{L}_{data} = \frac{1}{N}\sum_{i=1}^{N}(y_i - \hat{y}_i)^2 \tag{4.4}$$

$$\mathcal{L}_{physics} = \frac{1}{N}\sum_{i=1}^{N}\left(y_i - \hat{P}(x_i)\right)^2 \tag{4.5}$$

Where $N$ is the number of datapoints, $\lambda$ is a hyperparameter controlling the weight of the physics-based loss term, and $\hat{P}(x_i)$ is the physics-based term calculated using equation 4.6.

$$P(x) = \frac{Applied\ Load}{Alloy\ strength + \epsilon} \tag{4.6}$$

Where $P(x)$ is a simple representation of the physical relationship between stress and strain, assuming linear behavior under applied load and $\epsilon$ is a small constant for numerical stability ($\epsilon = 1 \times 10^{-7}$ in the present study). The physics term is further normalized depicted in equation 4.7 for the numerical stability and to match the scale of displacement values.

$$P'(x) = \frac{P(x) - \mu_p}{\sigma_p + \epsilon} \tag{4.7}$$

Where $\mu_p$ is the mean of $P(x)$ across the batch, and $\sigma_p$ is the standard deviation across the batch. This normalization ensures that the physics-based loss functions at the same scale as the data-driven loss.

The model is trained using the Adam optimizer with a learning rate of 0.001. A custom training step is constructed using TensorFlow's GradientTape, which simplifies the computation of gradients for the loss function in relation to the model's parameters. These gradients are then used to update the model weights via the optimizer, ensuring efficient convergence throughout the optimization process.

Figure 4.4 a) depicts a continuous displacement prediction across alloy strength and applied load combinations, with larger displacements (red) occurring in low strength-high load regions and smaller displacements (blue) in high strength-low load regions. Figure 4.4 b) shows the actual data points utilized for training, indicating that the model's predictions are consistent with experimental values. The discrete color-coded dots provide clear strength-dependent displacement behavior. Figure 4.4 b) shows rapid early convergence followed by steady refinement, with loss stabilizing at approximately 0.3 after 600 epochs, indicating effective model training.

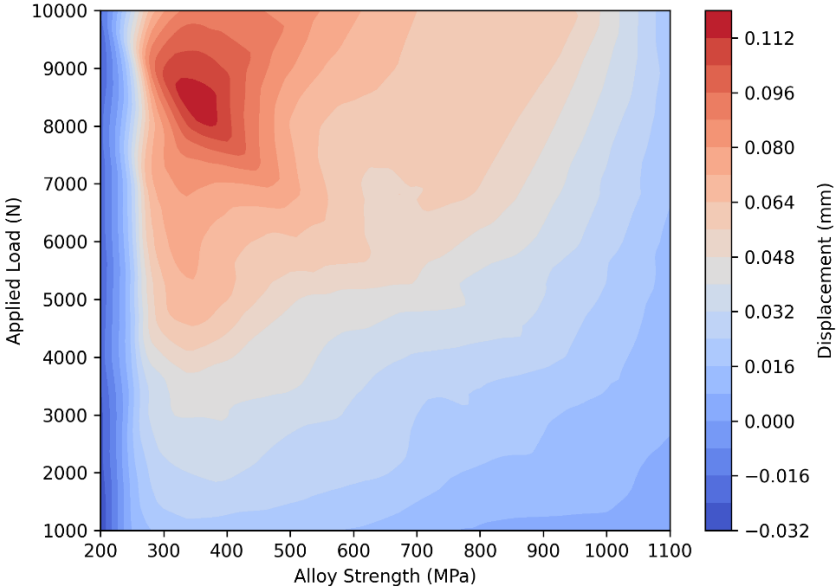

a)

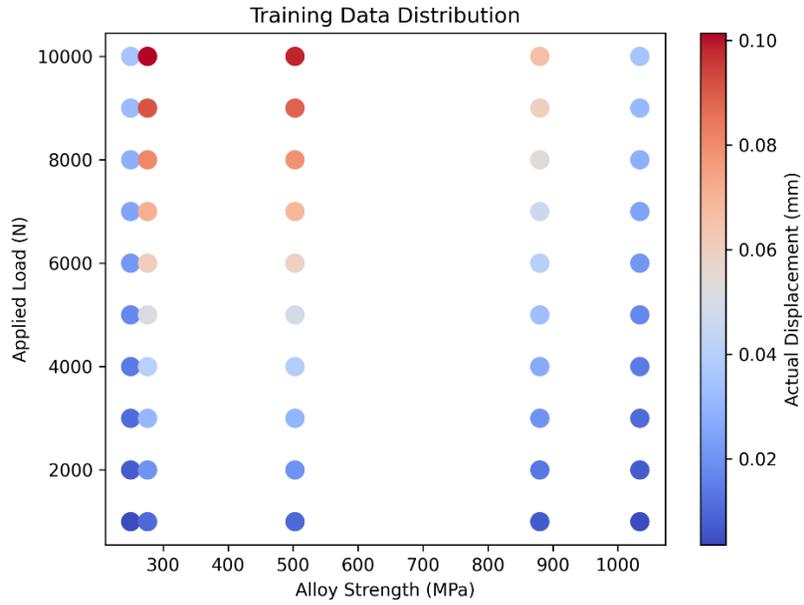

b)

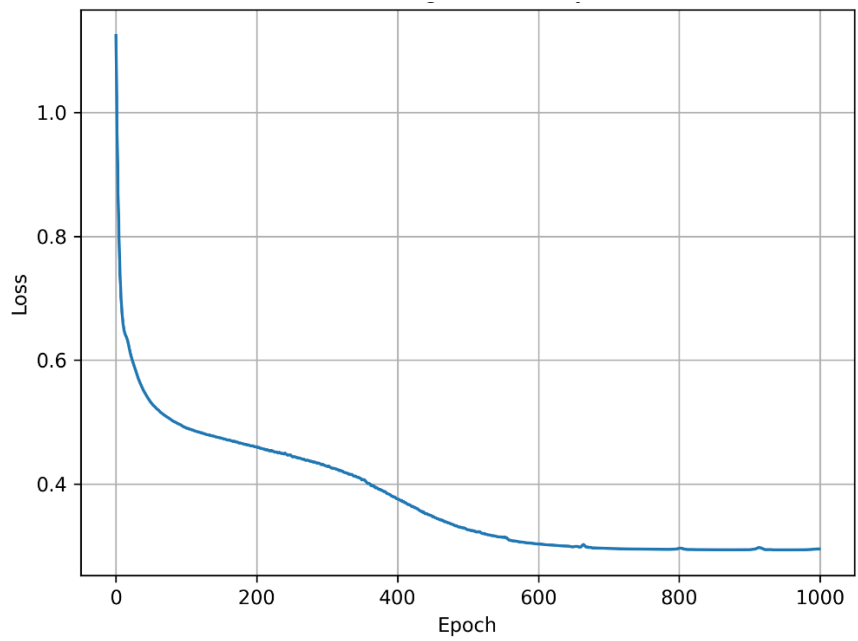

c)

**Figure 4.4.** Comprehensive visualization of PINN model performance showing a) predicted displacement surface across material strength and load ranges, b) actual training data distribution, and c) model convergence through training epochs. The color gradients represent displacement magnitude in millimeters.

The models performance is evaluated using three metric features i.e. $R^2$ score, Mean Square error (MSE), and Mean Absolute Error (MAE) shown in Figure 4.5. $R^2$ score depicted in equation 4.8 determines how well the model's predictions match the actual values. MSE calculates the average squared difference between the predicted and actual values as depicted in equation 4.9. MAE calculates the average absolute difference between predicted and actual values as depicted in equation 4.10.

$$R^2 = 1 - \frac{\sum_{i=1}^{N}(y_i - \hat{y}_i)^2}{\sum_{i=1}^{N}(y_i - \bar{y})^2} \quad (4.8)$$

$$MSE = \frac{1}{N}\sum_{i=1}^{N}(y_i - \hat{y}_i)^2 \quad (4.9)$$

$$MAE = \frac{1}{N}\sum_{i=1}^{N}|y_i - \hat{y}_i| \quad (4.10)$$

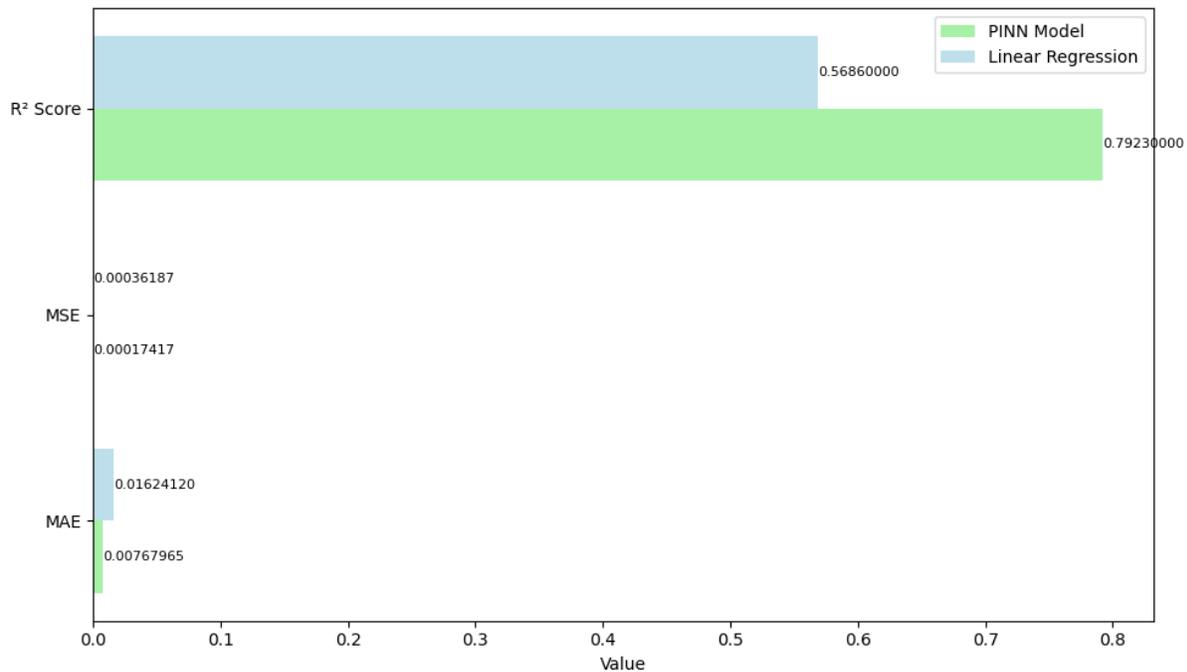

**Figure 4.5.** Comparison of metrics features obtained for linear regression model and PINN model

A comparison of performance metrics between the PINN and Linear Regression models indicates considerable variations in predicting ability as shown in Figure 4.1. The PINN model outperforms all important measures. The PINN model has a higher $R^2$ Score (0.7923) than Linear Regression (0.5686), indicating that it explains more of the dependent variable's variance. This significant difference of about 0.22 points shows that the PINN model reflects the data's underlying patterns more well. In terms of error metrics, the PINN model likewise performs significantly better. The PINN model's Mean Squared Error (MSE) is 0.00017417, which is less than half of the Linear Regression's MSE of 0.00036187. This lower MSE shows that the PINN model's predictions have lower average squared deviations from the true values. Similarly, the Mean Absolute Error (MAE) shows the same pattern, with PINN obtaining 0.00767965 versus Linear Regression's 0.01624120. This suggests that the PINN model's predictions depart less from true values in absolute terms, with an MAE about 52% lower than the Linear Regression model.

The distribution plots shown in Figure 4.6 of prediction errors for the Linear Regression and PINN models show major differences in their predictive tendencies. The Linear Regression model has a more symmetrical, bell-shaped error distribution centered at 0.01 mm, indicating a consistent but minor overestimation bias in its predictions. This symmetrical pattern suggests that the model's mistakes are uniformly distributed on both sides of the mean, as is common for linear models dealing with complex interactions. In comparison, the PINN model has a much different error distribution pattern, with a noticeable right-skewed shape. Its peak is closer to 0.00 mm, showing more accuracy in most predictions. The PINN distribution has a higher maximum density of around 37.0 than Linear Regression's 19.0, indicating that a greater proportion of its predictions cluster around true values. However, the PINN model's distribution has a broader right tail that extends to around 0.04 mm, showing that while it generally performs better, it may occasionally yield bigger errors in specific instances. The error ranges are also different across the two models, with Linear Regression covering from -0.02 mm to 0.03 mm in a more uniform spread, whilst the PINN model's faults range from about -0.01 mm to 0.04 mm. The PINN's sharper peak and concentrated distribution around zero error illustrate its superior predictive performance in the majority of cases, despite the presence of infrequent outliers. These distributional properties are consistent with and complement the preceding performance measures, demonstrating the PINN model's overall improved prediction accuracy over the standard Linear Regression technique.

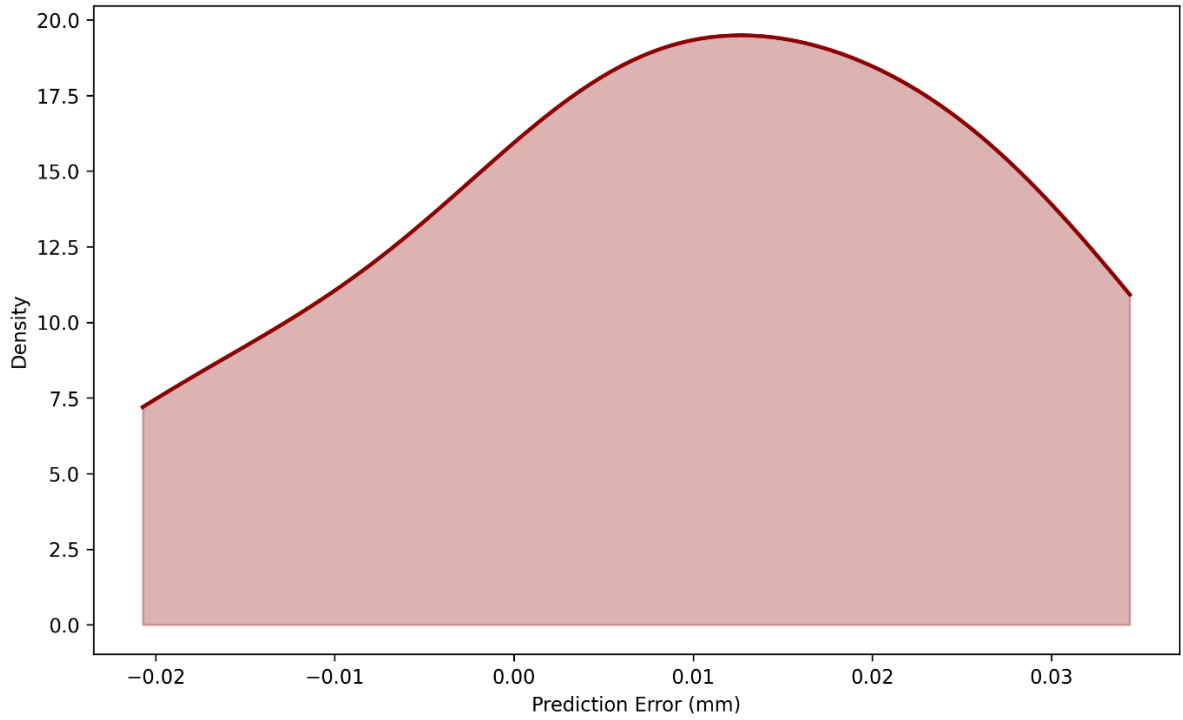

a)

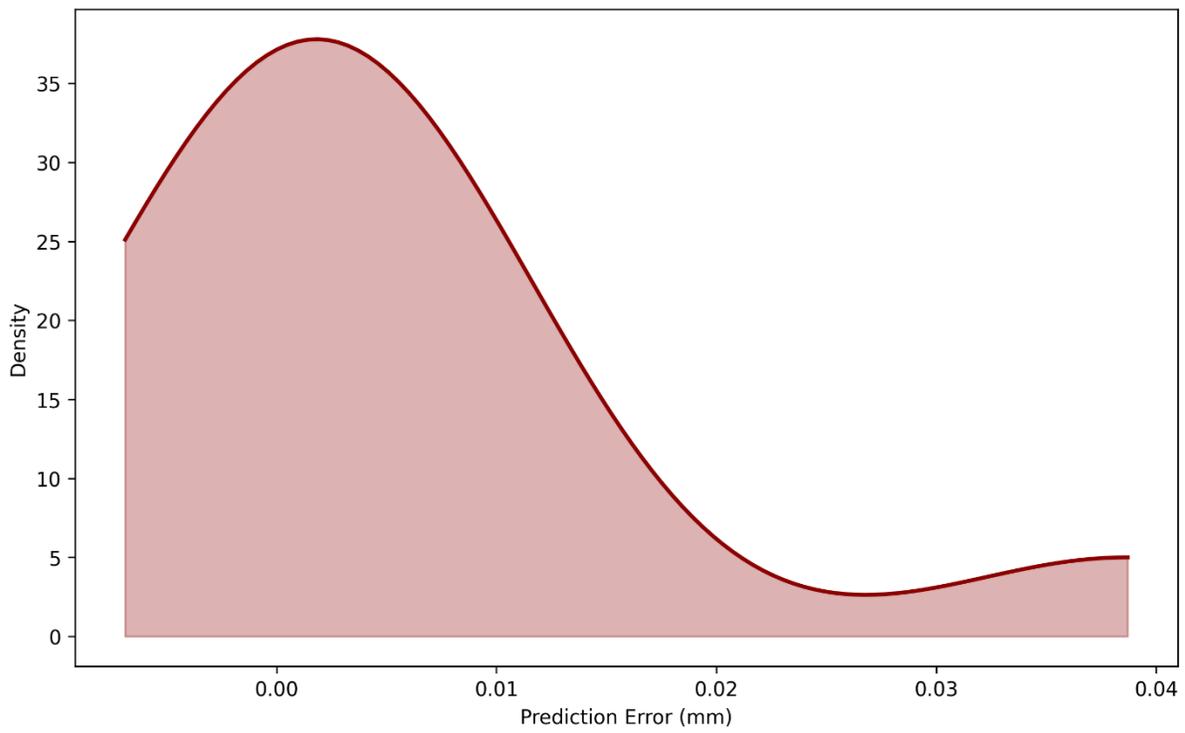

b)

**Figure 4.6.** Error distribution plot for a) Linear regression model, and b) PINN model

A comparison of actual versus predicted displacement plots for PINN and Linear Regression models, Figure 4.7a) and Figure 4.7b), respectively, indicates large differences in their predictive performances. The dashed line in the plots reflects perfect prediction, where actual and predicted values are equal, and colors for data points correspond to their magnitude of absolute error, ranging from blue-low error-to red-higher error. The PINN model gives a very good prediction accuracy and has most of the points tightly grouped around the perfect prediction line over the entire range of displacements between 0.00 and 0.10 mm. Most of the predictions are of relatively lower magnitudes of absolute errors represented by mostly blue-colored points but for a few at large values of displacements. This consistent clustering along the diagonal line indicates that the PINN model has successfully grasped the linearity and nonlinearity of the displacement relationship.

On the other side, the Linear Regression model gives more scattered predictions with a lot of deviation from the perfect prediction line, particularly for the middle range between 0.04 and 0.08 mm. The color gradient of the points shows a trend in increasing prediction errors with more points showing lighter blue to red colors, which indicates bigger absolute errors compared to the PINN model. A systematic deviation of points from the diagonal line, especially for the mid-range values, would mean that the Linear Regression model is not able to capture the underlying complexity of the relationship between displacements. It can also be seen from the visualization that both models have problems with the extreme values, mainly around the biggest displacement measures of 0.10 mm, where the prediction errors of both models increase, given by the red points.

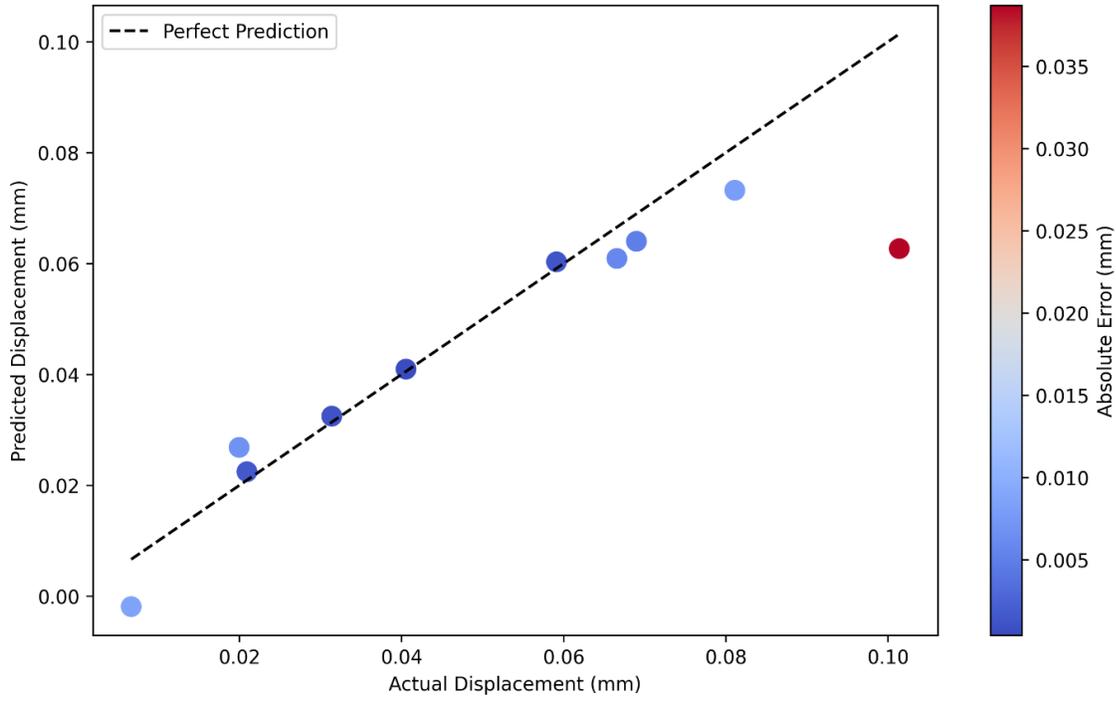

a)

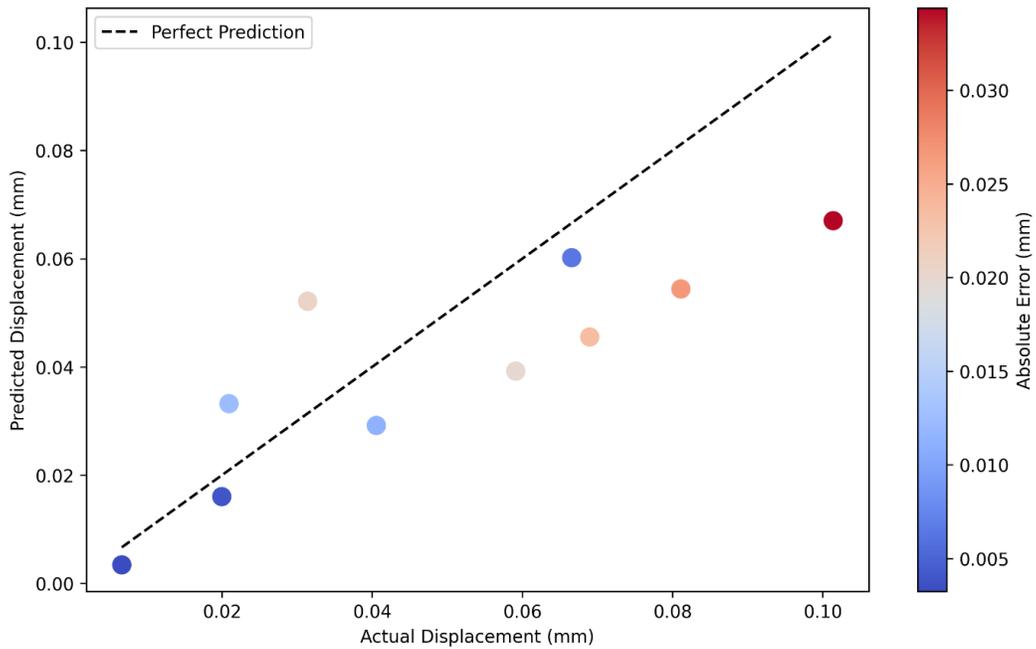

b)

**Figure 4.7.** Actual vs Predicted displacement magnitude plots for a) PINN model, and b) Linear regression model

In particular, residual plots for Linear Regression shown in Figure 4.8 a) and PINN depicted in Figure 4.8 b) show dramatically different patterns in their prediction errors for different displacement values. The residuals are the differences between predicted and actual values; the dashed line at zero represents a perfect prediction, while point colors show absolute magnitude errors ranging from blue (low) to red (high).

The residual plot of the Linear Regression model does show a pattern in the residuals that is somewhat troubling: it shows residuals increasing with larger actual values. Indeed, the residuals for smaller displacements—starting from 0.00 to 0.04 mm—stay relatively small and close to zero; this can be seen by the blue dots. However, there is an upward trend in residuals as the actual values increase, and the largest residuals (about 0.034) are at the highest level of displacement values (at 0.10 mm). Such a pattern, in the progression from blue to red of the color scheme, indicative of underpredicting larger displacement values, could be seen as an implication of a biased Linear Regression model. On the other hand, the residual plot for the PINN model shows more homogeneous and controlled error patterns. Residuals are generally smaller in magnitude, with most points clustered closer to the zero line and showing predominantly blue coloring, which indicates lower absolute errors. While there is still one noticeable outlier at the highest displacement value of 0.10 mm, the overall spread of residuals is more uniform across the range of actual values. The PINN model keeps the prediction accuracy relatively stable for different values of displacement without showing a systematic bias as was seen in the Linear Regression model. It also shows better balance in the residuals distribution above and below zero, which may hint at less biased predictions.

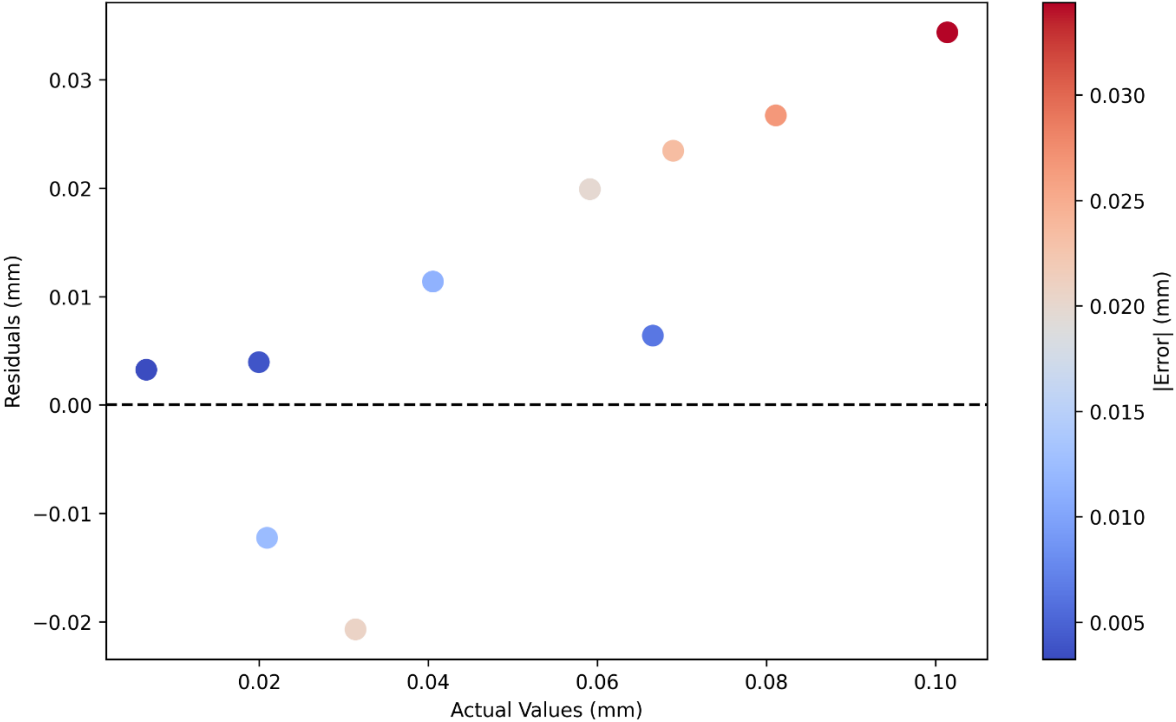

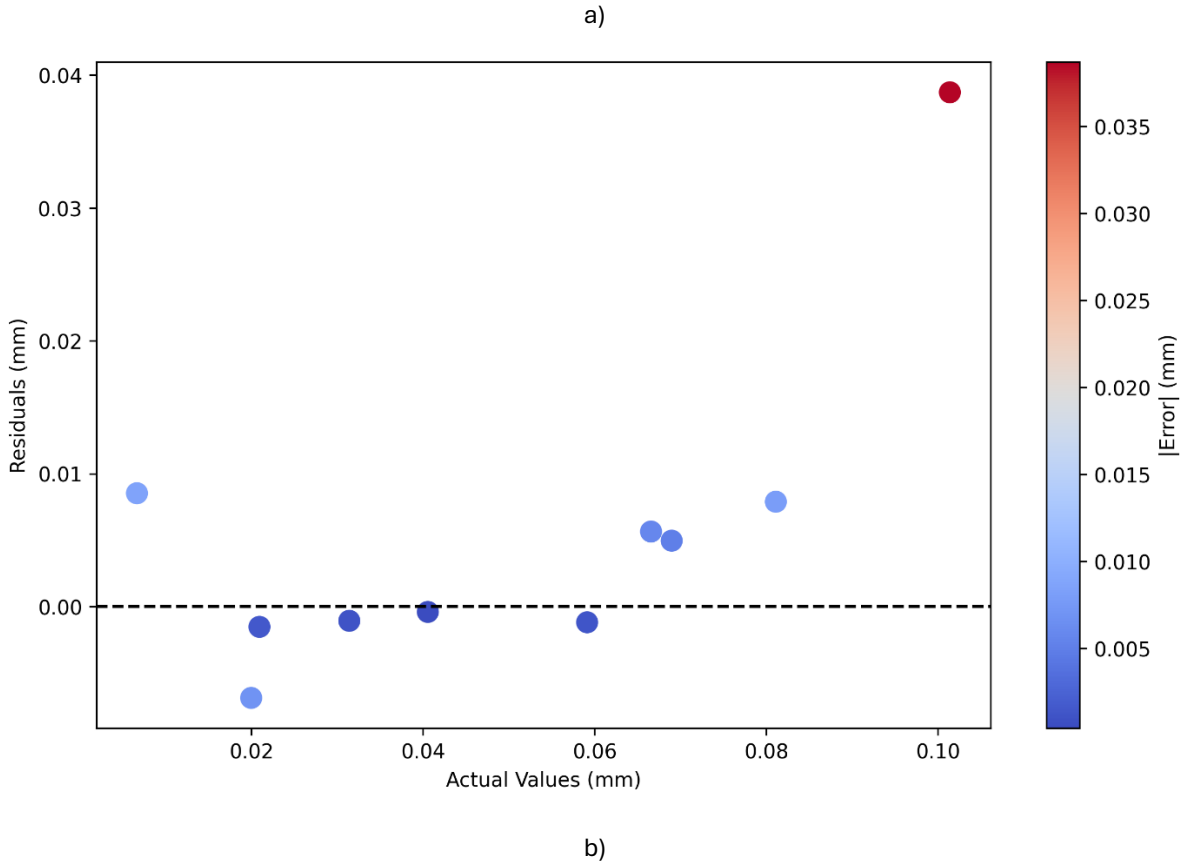

**Figure 4.8.** Residual plots for plot for a)Linear regression model, and b) PINN model

The 3D surface plots for the PINN shown in Figure 4.9 a) and Linear Regression shown in Figure 4.9 b) models show that there is a stark difference on how both models estimate displacement from alloy strength and applied load . Both of the visualizations map the predicted displacement values to a blue to red color scale to reflect the displacement level. The surface plot depicted for the PINN model reveals a more curved three-dimensional profile of the variables. It illustrates a curved shaped surface with different gradients, especially areas with larger applied load and smaller alloys. It appears that the PINN model has succeeded in capturing nuanced relationships between the strength of the alloy used and the applied load due to surface features such as gentle waves and a higher degree of rounding on the outer surface. These displacement values have an approximate order of 0.00 and 0.08 mm, and the maximum displacements (depicted in red) are observed at high loads and low alloy strength. However, the Linear Regression model gives a significantly more straightforward plane with equal slopes in the entirety of the prediction space. This means that this linear surface represents a direct relationship meaning displacement rises uniformly as load goes up and as alloy strength goes down. The displacement range is smaller: 0.01-0.06 mm and the blending from a blue to a red colour is gradual to highly predictable. This linear behavior is an inherent problem with the model since it can only account for first order effects between the variables.

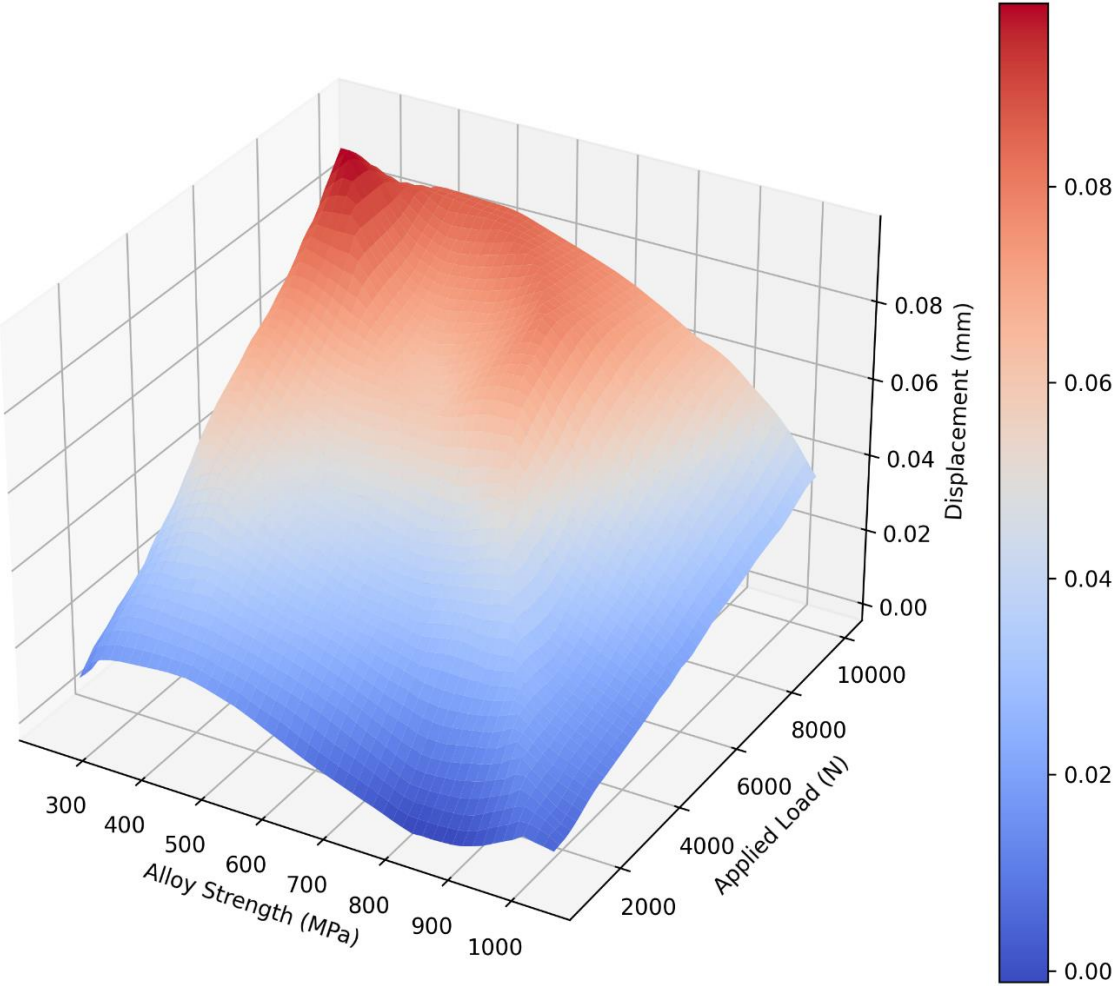

a)

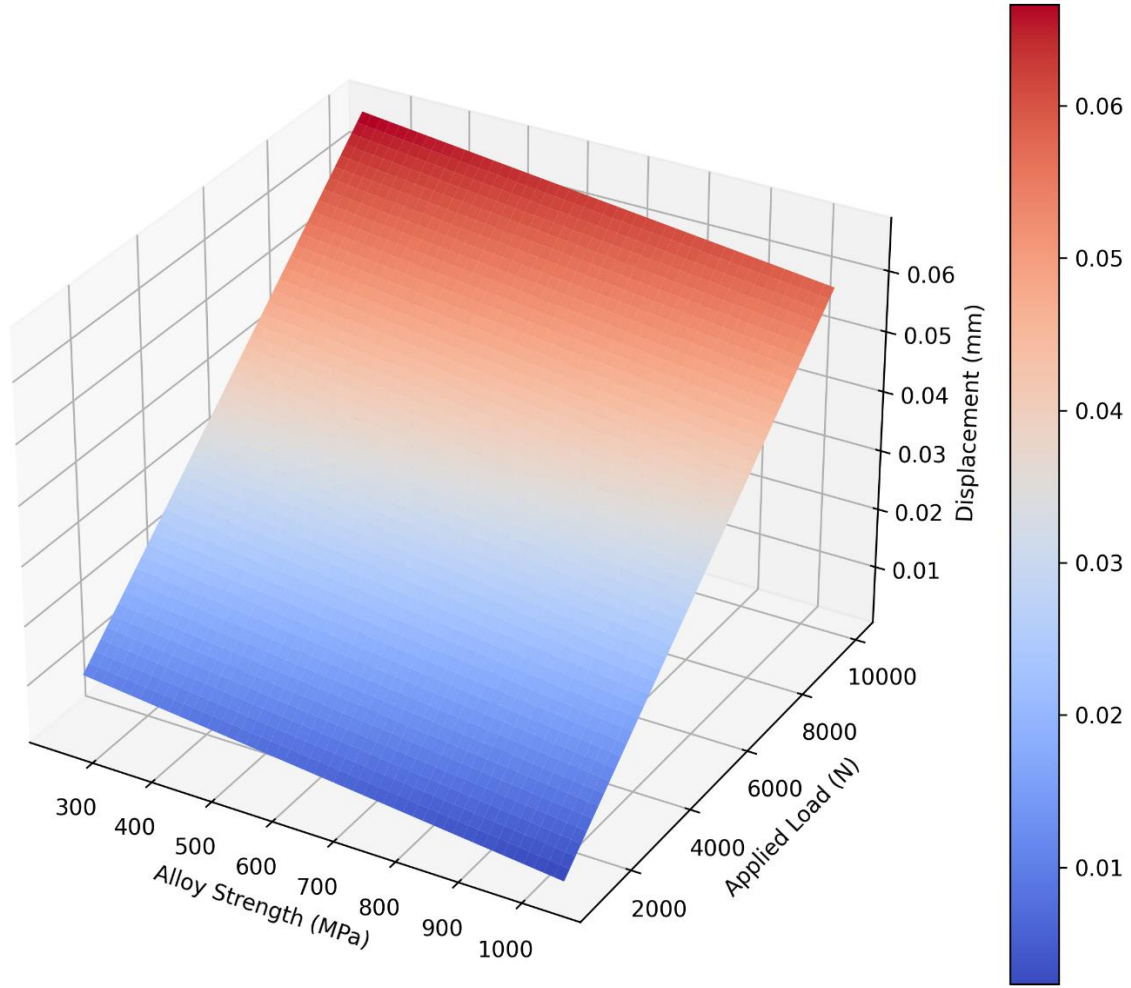

b)

**Figure 4.9.** 3D surface visualization of predicted displacement for a) PINN model, and b) Linear regression model

## 5. Conclusion

This study shows that Physics-Informed Neural Networks can accurately anticipate the mechanical behavior of architected lattice structures. The proposed PINN model outperformed traditional linear regression, obtaining 39% higher accuracy in $R^2$ score and lowering prediction errors by almost 50%. The model successfully reflects the complicated interactions between material properties, applied loads, and consequent deformations, and it excels at handling non-linear behaviors that conventional techniques struggle with.

Future research directions could include:

- Extending the model to anticipate dynamic loading and fatigue behavior.
- Incorporating microstructural characteristics and manufacturing restrictions
- Creating real-time optimization tools for lattice structure design
- Increasing the material database to include composites and functionally graded materials.
- Using multi-objective optimization for both strength and weight considerations
- Examining thermal-mechanical coupling effects in lattice structures